\documentclass{ieeeaccess}
\usepackage{amsmath,amssymb,amsfonts}
\usepackage{algorithmic}
\usepackage{graphicx}
\usepackage{textcomp}
\usepackage{lscape}
\usepackage{float}
\usepackage{caption}
\usepackage[sorting=none]{biblatex}
\def\BibTeX{{\rm B\kern-.05em{\sc i\kern-.025em b}\kern-.08em
T\kern-.1667em\lower.7ex\hbox{E}\kern-.125emX}}

\begin{document}

\history{2022-03-09, MM805 Literature Review}
\doi{None}

\title{Image Style Transfer: from Artistic to Photorealistic}

\author{Chenggui Sun and Li Bin Song}
\author{\uppercase{Chenggui Sun}\authorrefmark{1}
AND \uppercase{Li Bin Song\authorrefmark{2}}
}
\address[1]{Department of Computing Science, University of Alberta, Edmonton, Alberta, Canada T6G 2E8 (e-mail: chenggui@ualberta.ca)}
\address[2]{Department of Computing Science, University of Alberta, Edmonton, Alberta, Canada T6G 2E8 (e-mail: libin3@ualberta.ca)}


\corresp{Corresponding author: Chenggui Sun (e-mail: chenggui@ualberta.ca) and Li Bin Song(libin3@ualberta.ca)}




\titlepgskip=-15pt
\maketitle

\section{Summary}

The rapid advancement of deep learning has significantly boomed the development of photorealistic style transfer. In this review, we reviewed the development of photorealistic style transfer starting from artistic style transfer and the contribution of traditional image processing techniques on photorealistic style transfer, including some work that had been completed in the Multimedia lab at the University of Alberta. Many techniques were discussed in this review. However, our focus is on VGG-based techniques, whitening and coloring transform (WCTs) based techniques, the combination of deep learning with traditional image processing techniques.

\section{Literature Review}

\subsection{Computer Vision and Style Transfer}
Computer vision has been playing more and more important role in every aspect of our society, including medical \cite{cheng2007airway, rossol2011framework, faraji2018segmentation, xu2009gradient, mukherjee2010automatic}, education \cite{cheng2009interactive}, communication \cite{basu1998enhancing, cheng2007perceptually, baldwin1999panoramic, singh2005visual, basu1993variable, basu1994videoconferencing},  identification verification \cite{yin2001generating, yin2001nose, yin1999integrating, berretti2018representation, singh2005pose, rossol2015multisensor}, object tracking \cite{singh2005gaussian}. Image style transfer or stylization is one of the computer vision domains that have been attracting more and more research attention. It extracts features from one or more images specified as the reference style and performs stylization on an input image specified as the content image whose structure is kept intact. As a result, it produces an output image combining the content of the input image and the style of a reference image or images \cite{liu2019advanced}. There are two types of image style transfers: artistic and photorealistic style transfer. Photorealistic style transfer synthesizes realistic images from an input image specified as the content image with a photo specified as the reference image \cite{zhao2020survey}. 

\begin{figure}
    \centering
    \includegraphics[width=0.8\columnwidth]{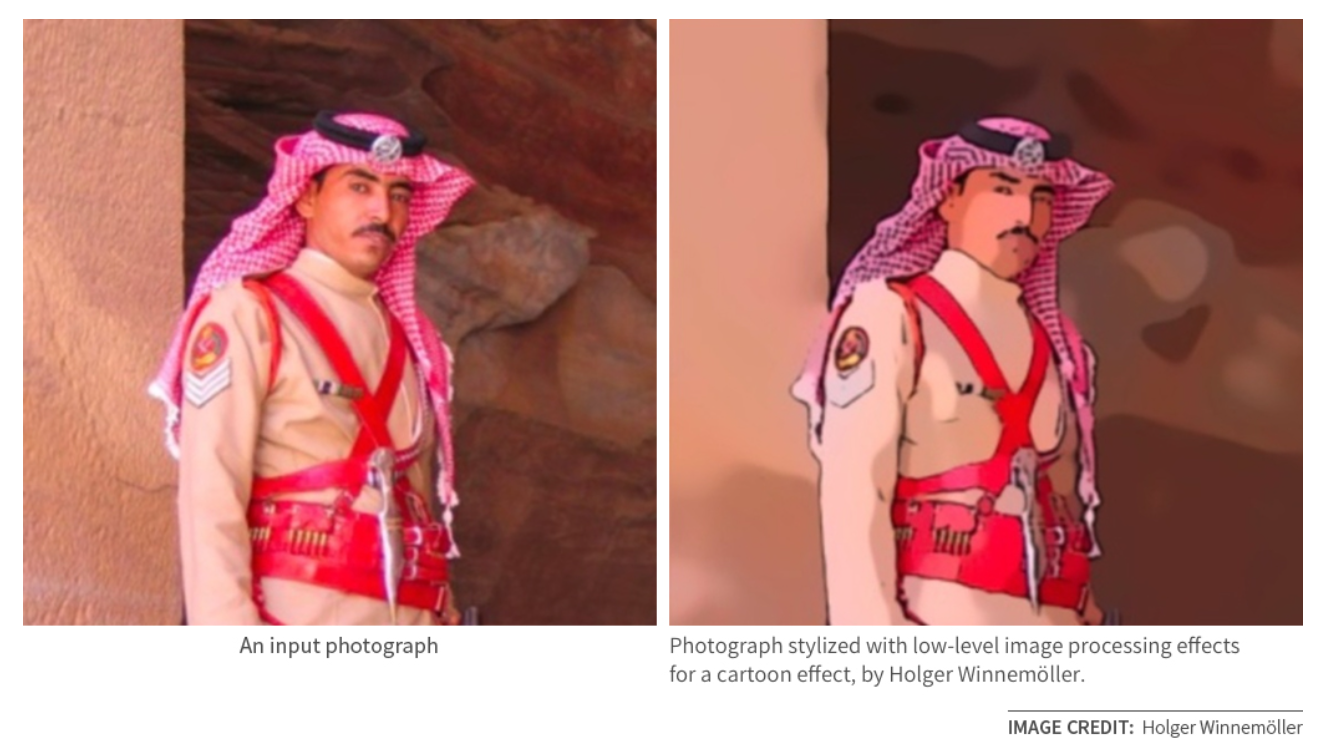}
    \caption{Style transfer filters \cite{hertzmann_2018}}
    \label{fig:filter1}
\end{figure}

The image style transfer methods were used to be called texture transfer methods. During the past several years, inspired by the rapid development of neural network algorithms, the performance of image style transfer has been significantly improved \cite{liu2019advanced}. The original style transfer idea was to create artistic images from a photo. The early algorithms are based on image filter functions \cite{hertzmann_2018}. Holger Winnemoeller presented a Real-time video abstraction method in 2006 \cite{real-time_abstraction}. This method shows that the essential computer image filter functions, blurring and sharpening can create artistic cartoon-like effects, as shown in Figure \ref{fig:filter1}. 


Many other algorithms were developed later to simulate artists' techniques, such as brush strokes, or pen strokes. These techniques can create effects that look like images were painted or drawn by artists. All these algorithms were developed on top of classical computer vision image filters. And each of these algorithms requires coding to get a specific artistic style.

Patch-Based Synthesis is a next-level style transfer method. This method comes from the question of whether the computer can recognize textures. A human can do the job very efficiently. He/she can realize the texture pattern even if he/she sees the texture pattern for the first time. But for computers, it is not that easy. The patch-based method divides the challenge into two tasks. The first task is to recognize the texture pattern, and the second task is to apply the pattern to the target image. A straightforward way to resolve this problem is to create a random new image with the same statistics information from the style image. 

An alternative image patch-based method was created later. This method uses a square of pixels in an image, e.g. 5x5, 19x19. The assumption is that if this small patch is texture, we can find a similar path on other pictures. The procedure to generate texture synthesis images with this model is like this: 1) create an empty image, 2) fill in pixels one by one, 3) randomly sample new pixels from the input texture(patches), 4) keep the pixels if we can find similar pixels from input patches \cite{hertzmann_2018_2}.

\begin{figure}[h]
\centering
\includegraphics[width=0.8\columnwidth]{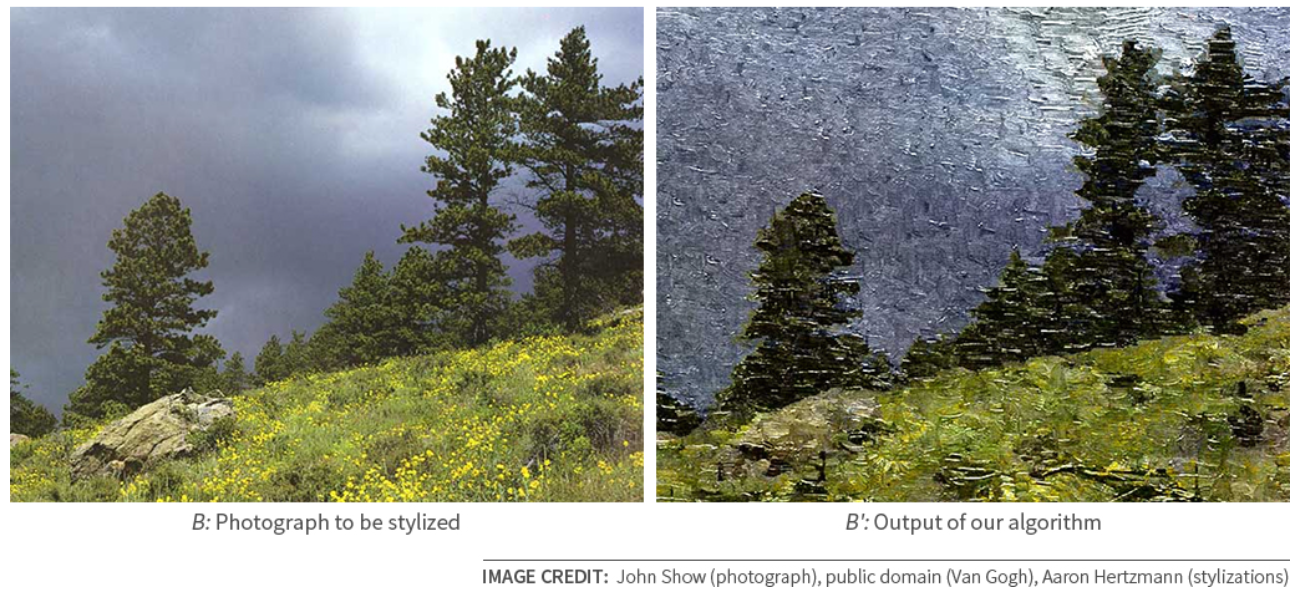}
\caption{patch based style transfer \cite{hertzmann_2018_2} }
\label{fig:patch1}
\end{figure}


\subsection{Artistic Style Transfer}

\begin{figure}[h]
\centering
\includegraphics[width=0.8\columnwidth]{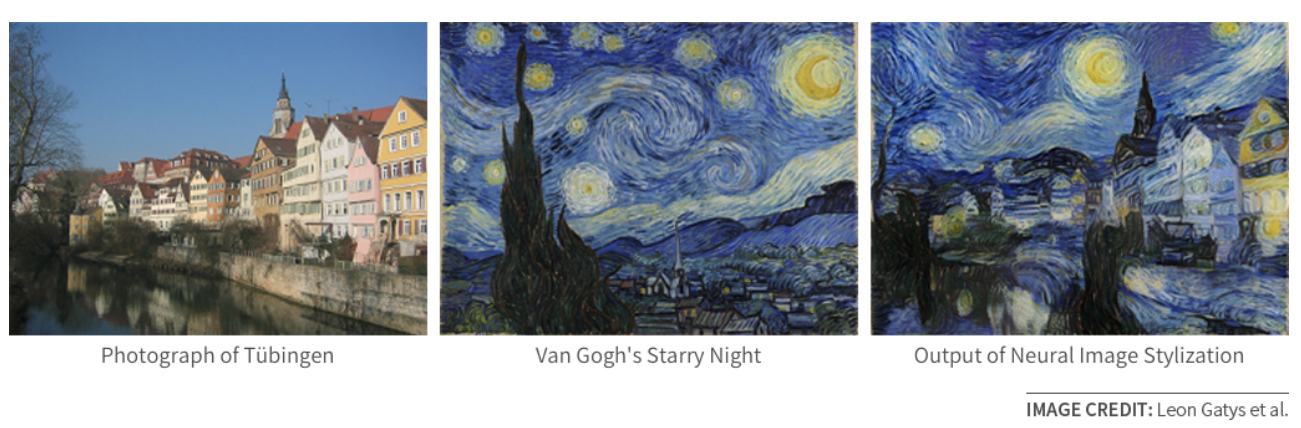}
\caption{Neural style transfer \cite{gatys2016image}}
\label{fig:neuralstyle}
\end{figure}

Neural networks and machine learning achieved huge success on the object recognition  challenge after 2012. It’s reported that the first effective and complete neural method to perform artistic and photorealistic style transfer was proposed by Gatys et al. \cite{zhao2020survey}. In their work, the authors proposed that high-performing Convolutional Neural Networks can learn the generic feature representations of content and reference images independently, then manipulated the output images with the content from the content images and the styles of reference images. They introduced a neural network-based artistic stylization method, which is a texture transfer algorithm that synthesized texture using feature representations from state-of-the-art Convolutional Neural Networks. They used the feature space provided by a 19-layer VGG network consisting of 16 convolutional and 5 pooling layers. To keep the mean activation of each convolutional filter over images and positions to be one, they scaled the weights to normalize the VGG network. The scaling for the VGG network didn’t change its output, because its activation function is rectifying linear activation functions and it didn’t contain normalization or pooling over feature maps. For synthesizing images, they found that it yielded more appealing results to replace the maximum pooling operation with average pooling. While their algorithms synthesized high perceptual quality images, it still has some technical limitations. The most limiting factor is very visible: the resolution of the synthesized images. Because the dimensionality of the optimization and the number of units in the Convolutional Neural Network is linearly related to the number of pixels, the image resolution has a huge influence on the synthesis speed. They synthesized images on an Nvidia K40 GPU, which could take up to one hour. The exact time span of the synthesis procedure depended on the image size and the stopping criteria for the gradient descent.  Another issue with their algorithm is the subjection of the synthesized images to some low-level noise. This is less of an issue when they performed artistic style transfer. However, this problem became more significant for photorealistic style transfer. The photorealism of the synthesized image was very poor. One possible solution to mitigate this problem is to construct efficient de-noising techniques to post-process the images or pre-process the extracted features of reference photos. Many researchers had done excellent work in this area, which will be introduced later in this review \cite{gatys2016image}. This method delivered awe-inspiring results as shown in Figure \ref{fig:neuralstyle}.

Since Gatys’s work, many researchers have published their work to address different aspects of style transfer including speed, quality, user control, diversity, semantics understanding, and photorealism. However, these methods are lack generalization in neural network design \cite{li2017universal}. Li et al. proposed a method that can achieve universal style transfer \cite{li2017universal}. In the transfer process, the authors applied the classic signal whitening and coloring transforms (WCTs) on the content features at intermediate layers to transform them to have the same statistical characteristics of the style features at the same layer. Their method not only can conduct the single-level stylization, but also the multi-level stylization. The multi-level stylization can synthesize images with comparable or even better visual quality with much less computational costs. The multi-level stylization is completed by applying WCT sequentially to multiple feature layers. The steps to apply WCT to a single layer is listed as follows, which is also how the single-level stylization is completed:

\begin{itemize}
    \item The VGG-19 network was employed as the encoder, and an asymmetric decoder was trained to reconstruct the original image by inverting the VGG-19 features. The trained encoder and the decoder were fixed for all image style transfers in the reported research; 
    \item WCT was applied to transform one layer of content features to make its covariance matrix to match the covariance matrix of the style features of the same layer. Therefore, the content features were transformed and the style transfer was completed; 
    \item Then, the downstream decoder was utilized to reconstruct the stylized images from the transformed features.
\end{itemize}

When given a new style, this algorithm only needs to extract corresponding feature covariance matrices and apply them with WCT to transform the content features, which is how this algorithm achieves universal style transfer. The degree of style transfer aka the balance between stylization and content preservation can be defined by the users using a control parameter that comes with this algorithm \cite{li2017universal}.

\begin{figure}[h]
\centering
\includegraphics[width=0.8\columnwidth]{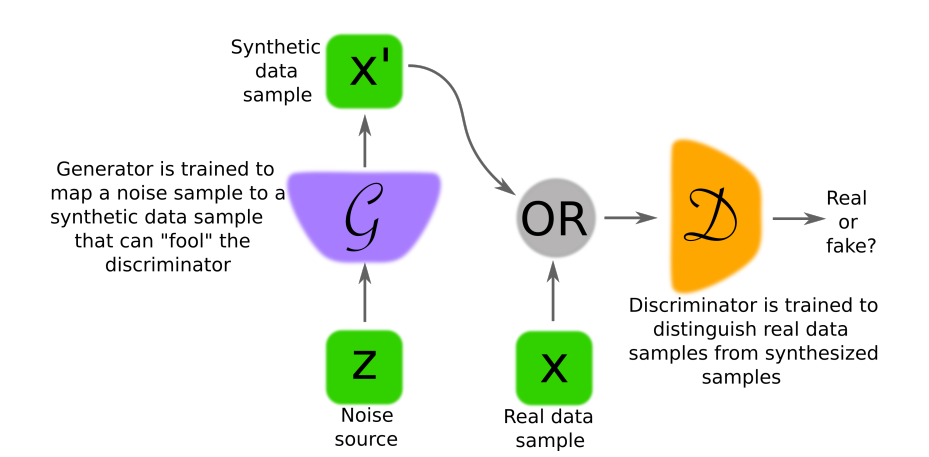}
\caption{"GAN" network \cite{GAN_Overview}}
\label{fig:GAN1}
\end{figure}

Generative Adversarial Network(GAN) is a semi-supervised and unsupervised machine learning network framework. Ian Goodfellow’s team proposes it in 2014. The key idea is to train two models at the same time: a generative model G and a discriminative model(D) \cite{GAN_Intro}. A high-level GAN design is shown in Figure\ref{fig:GAN1}. The generator (G) does not access the actual data; it uses noise source data to generate synthetic data samples. Then, these samples passes to the OR gate, which mixed the input with existing data. The Discriminator takes the output from the OR gate and justice if it is real or fake. Both G and D are trained in the training process until D cannot tell if the sample is real or fake. At that time, we got an excellent Generator that we could be used to generate synthetic data \cite{GAN_Overview}.

\begin{figure}[h]
\centering
\includegraphics[width=0.8\columnwidth]{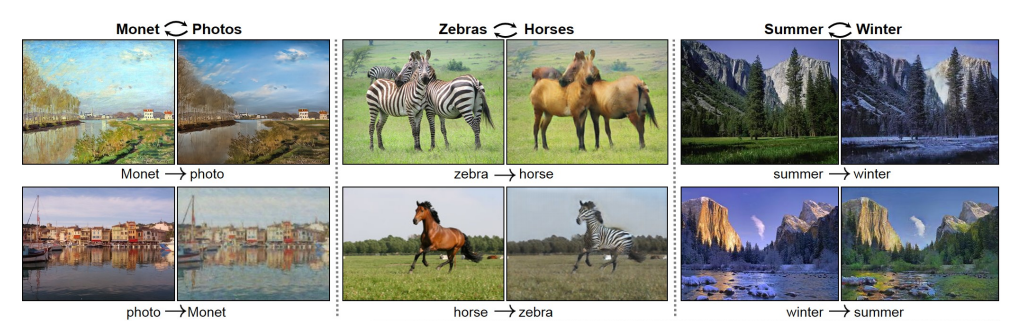}
\caption{Image generated CycleGAN \cite{GAN_Overview}}
\label{fig:GAN2}
\end{figure}

Once we get a trained GAN network, the components G and D can be reused for other purposes. For example, the outputs of the convolution layers of the model D can be used as a feature extractor. By adding linear models fitted on top of the feature extractor layer, it can be used for classification tasks. Style transfer is one of the most popular applications of GANs. Examples of popular applications include converting a photo to artistic style, converting a summer photo to a winter photo, converting a horse to a zebra. An example is shown in Figure \ref{fig:GAN2}. 

There are also challenges for GAN networks, such as Model Collapse and training instability\cite{GAN_Overview}. As a result, more GAN networks are designed for specific use domains.

\subsection{2D and 3D Photorealistic Style Transfer}

When the Neural Network algorithm of Gatys et al. applied for photorealistic style transfer, it produces images with distortions. Building upon the research done by Gatys et al., Luan et al. \cite{luan2017deep} proposed an approach to augment the Neural Style algorithm proposed by Gatys et al. by introducing a photorealism regularization term in the objective function to reduce distortions and an optional guidance to the style transfer process to mitigate the content-mismatching. The details about their regularization and optional guidance are:
\begin{itemize}
    \item Photorealism regularization: it regularizes the optimization scheme of style transfer to preserve the structure of the input image and produce photorealistic outputs. The authors express the regularization constraint on the transformation applied to the input image to penalize image distortions. This is achieved by seeking an image transform that is locally affine in color space, which is an affine function mapping the RGB values of the input image onto their output counterparts. Mathematically, the author proposed the regularization built upon the Matting Laplacian that describes a least-squares penalty function that can be minimized with a standard linear system represented by a matrix of the input image. They used the vectorized version of the output image to define the regularization term to penalize outputs that are not well explained by a locally affine combination of the input RGB channels.
    \item Optional guidance to the style transfer process based on semantic segmentation of the inputs: Gatys’ algorithm implicitly encoded the exact distribution of neural responses, and wasn’t able to adapt to variations of semantic context. The authors addressed this problem by generating image segmentation masks for the common labels of the input and reference images, such as sky, buildings, water, etc. The masks added to the input images were treated as additional channels. These segmentation channels were concatenated to augment the style loss. The labels of the reference image were assigned to the input semantic labels to avoid the occurrence of “orphan semantic labels”. The assigned labels are generally equivalent in the context, for example, “lake” was treated as the same as “sea”. 
\end{itemize}

Although the method proposed by Luan et al. solved the optimization problem and improved the quality of synthesized images, it still generates stylization with noticeable artifacts and its computational costs are very high, which limits its use in practical scenarios \cite{yoo2019photorealistic}. Buiding up the work of Luan et ail, Li et al. proposed a method called PhotoWCT \cite{li2018closed}. Their photorealistic stylization algorithm consists of a stylization transform step and a photorealistic smoothing step:
\begin{itemize}
    \item Stylization: inspired by the WCT method of Luan et al. and the unpooling layer’s capability in preserving spatial information, the authors replace the upsampling of the decoder with unpooling layers.
    \item Smoothing: because the inconsistent stylization of the semantically similar regions still exists, the authors defined an affine matrix to describe the pixel similarity of the content in a local neighborhood, and employed the affinities in the content photo to facilitate the pixels in a local neighborhood to be stylized similarly aka the smoothing of the stylized result.
\end{itemize}

The photoWCT method improved the results of stylization. However, its post-process steps require hefty computation power and time, and to set hyperparameters manually \cite{yoo2019photorealistic}. Yoo et al. proposed a wavelet corrected transfer based on WCT, which substituted the pooling and unpooling in the encoder and decoder with wavelet pooling and unpooling \cite{yoo2019photorealistic}. Attributing to wavelets’ properties of providing minimal information loss, this architecture change empowers WCT to fully reconstruct the images without any post-processes. Yoo’s method is also integrated with progressive stylization. As had been discussed earlier, WCT and PhotoWCT transformed features following a multi-level strategy. This wavelet corrected transfer progressively transforms features with a single pass. The details about the application wavelet transform in their algorithm and progressive transform are explained as follows:
\begin{itemize}
    \item Model architecture: the ImageNet pre-trained VGG-19 network was employed as the encoder. The decoder’s structure is the mirror of the encoder structure. The high-frequency components (LH, HL, HH) are passed to the decoder directly from the first wavelet pooling layer, where the low-frequency component (LL) is sent to the next encoding layer then the decoder. All components were aggregated at the mirrored wavelet unpooling layer of the first wavelet pooling layer.
    \item Progressive stylization: features were progressively transformed within a single forward-pass, where WCT was applied at each scale. WCT conducted style transfer by directly matching the correlation between the content features and style features. More WCTs can be added on skip-connections and decoding layers to strengthen the stylization when the increase of the time consumption is accepted.
\end{itemize}

Besides these classic articles regarding the photorealistic style transfer that were reviewed here, there are numerous works in style transfer that have been done. Mechrez et al. proposed a photorealistic style transfer that relies on the Screened Poisson Equation \cite{mechrez2017photorealistic}.  Wang et al. proposed an approach that consists of a loss network and a style fusion model. The loss network is a dual-stream deep convolution network, and the style fusion model is an edge-preserving filter. The novelty of this approach is that the authors introduced an additional similarity loss function to constrain the detail reconstruction and style transfer procedures \cite{luan2017deep}. Ding et al. presented an approach utilizing a deep neural network with an attention mechanism and wavelet transformation for decoupled style and detail synthesis \cite{ding2022deep}. Chiu and Gurari proposed a method called PhotoWCT2. This method was equipped with blockwise training to perform multi-level feature transformations and employed with skip connections of high-frequency residuals to preserve image quality when applying the sequential multi-level feature transformations \cite{chiu2022photowct2}. Xia et al. presented an algorithm that can transfer artistic styles of an image onto local regions of a target video while preserving photorealistic effects \cite{xia2021real}. Qu et al. constrained a non-local representation scheme with a mutual affine-transfer network to address photorealistic style transfer \cite{qu2021non}. Qiao et al. proposed a scheme constrained with a Style-Corpus Constrained Learning (SCCL) \cite{qiao2021efficient}. 

\begin{figure}[h]
\centering
\includegraphics[scale=0.4]{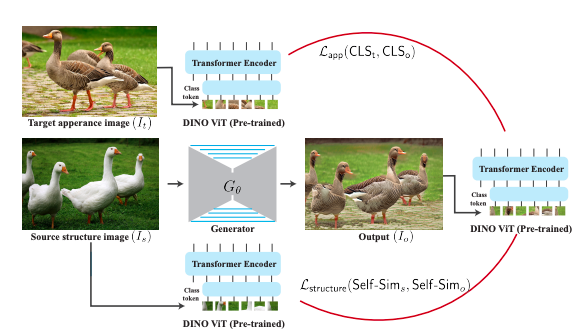}
\caption{ViT pipleline \cite{Splice2022}}
\label{fig:vit1}
\end{figure}

Recently, the Transformer that has achieved significant progress in the Natural Languages Process field,  is also introduced into computer vision tasks. Narek Tumanyan introduced a visual transformer for style transfer tasks between two photos from nature\cite{Splice2022}. The idea of this method is to transfer a natural photo's style to a content image and maintain the structure in the content images. The logic pipeline of ViT is shown in Figure \ref{fig:vit1}. In this pipeline, the critical component is the pre-trained DINO-ViT, a Vision Transformer. This ViT transformer is used to extract structure features from content and style images. After the structure and style information is extracted from the ViT features, the process will separate them from the self-attention modules. And the process converts structure into the self-similarity keys. After that, the process trains a generator to create a target output image that contains content image structure but with the texture from the style image.

The benefits of this method are:
\begin{itemize}
    \item Using pre-trained DINO-ViT transformer. 
    \item Single pair of image input. It requires a content image and a style image. This makes the method easy to adopt for various use cases.
    \item It preserves structure in a granular manner that fits photorealistic style transfer better.
\end{itemize}

One more aspect of style transfer that is worthy of detailed review is the 3D photorealistic style transfer, which spurred a soaring interest for its potential in the applications of mobile photography, augmented reality (AR), and virtual reality (VR) \cite{mu20213d}. Garcia et al. presented a coherent stylized rendering technique to stylize 3D scenes at the compositing stage. This method uses an implicit grid, which is based on the local object positions stored in a G-buffer, to generate a distribution of 3D anchor points that follow object motion \cite{garcia2021coherent}. Mu et al. proposed a deep model that can create 3D content from 2D images by learning geometry-aware content features from a point cloud representation of the scene to stylize images consistently across views \cite{mu20213d}. Their model performs 3D photo stylization following the following steps:
\begin{itemize}
    \item Use an input content images’ estimated depth map to back-project the image into an RGB point cloud; 
    \item “Inpaint” the point cloud to cover the scene’s dissoccluded parts then “normalized” the point cloud; 
    \item Design an efficient graph convolutional network to process the point cloud and extract 3D point-wise features for 3D stylization; 
    \item Adapt a style transfer module to modulate point-wise features with the reference style image; 
    \item Project the featurized points to novel views to stylize images consistently across views with a differentiable rasterizer.
\end{itemize}

\subsection{Traditional Methods and Style Transfer}

Although the application of deep learning neural networks in computer vision have significantly improved the capability of style transfer techniques, they still need to work with traditional methods to overcome their shortcoming in utilizing the generated features. Traditional techniques such as Fourier transform, Hough transform, Radon Transform, and Wavelet transform have been used to deal with computer vision problems associated with image feature processing before the booming of the deep learning era. 

Fiala and Basu reported that the Panoramic Hough transform has good potential to detected the distorted horizontal lines to supplement the trivial detection of vertical lines in panoramic imagery \cite{fiala2002hough}. The good grouping of Panoramic Hough transform peaks didn’t need preprocessing such as noise-reducing smoothing, and the horizontal line could be detected without advanced peak detection techniques or clustering algorithms because the parameter groupings in the parameter space were distinct enough \cite{fiala2002hough}. Hough Transform was also used to detect facial features with deformable template coupled with color information \cite{yin1999integrating}. Yin and Basu proposed a method with designed individual templates to extract the shape of the nostril and nose-side, and used the extracted feature shapes to guide a facial model for synthesizing realistic facial expressions \cite{yin2001nose}.  Singh et al. reported to extract pose information from the parametric Radon transform, which generated maximum correspondence to the specific orientations of the skeletal representation \cite{singh2005pose}. 

The applications of wavelet transform in style transfer have been reviewed in the previous section \cite{yoo2019photorealistic}. The implementation of wavelet transform can be treated as the implementation of a filter whose response is the desired wavelet function \cite{akansu2010emerging}. The Wavelet Transform of a signal x(t) can be described by \cite{chakrabarti1996architectures}:

\[
\hat{w}(u,s) = \int\hat{x}(t) \hat{h}(\frac{t-u}{s})dt
\]
where h(t) is the wavelet function. The Wavelet Transform of a sequence x(i) aka samples of a continuous signal spaced in time (b) and scale (a) is given by the following equation \cite{chakrabarti1996architectures}:
\[
W(b,a) =\frac{1}{{\left | a \right |}^{^0.5}}\sum_{i=b}^{i=al+b-1}x(i)h(\frac{i-b}{a})
\]
where L is the size of the basic wavelet's support. One type of wavelet transforms is Discrete Wavelet Transform (DWT), which is widely used in image processing. DWT can be treated as the multiresolution decomposition of a sequence. Its input is a sequence x(n) with a length N and output a length N sequence, which is the multiresolution representation of x(n). The implementation of 1 D DWT can follow the Pyramid Algorithm and the implementation of the separable 2 D DWT can be solved by the horizontal and vertical dimensional decomposition of 1 D DWT \cite{chakrabarti1996architectures}.

Williams and Li proposed Wavelet Pooling to the convolutional neural network as an alternative to traditional max pooling and average pooling \cite{williams2018wavelet}. Their method uses a second-level wavelet decomposition to subsample features and mitigate the overfitting problem that occurred with max-pooling by reducing features in a structurally compact manner. It needs to conduct a 2nd order decomposition in the wavelet domain following the fast wavelet transform (FWT) to perform his wavelet pooling. After the 2nd order decomposition, the image features are reconstructed using the 2nd order wavelet subbands based on the inverse DWT (IDWT):
\[
W_{\varphi}\left [j,k  \right ] = h_{\varphi}\left [-n  \right ]*W_{\varphi}\left [j+1,n  \right ] + h_{\psi }\left [-n  \right ]*W_{\psi}\left [j+1,n  \right ]|_{^{_{a=\frac{k}{2}, k\leq 0}}}
\]

The forward propagation of wavelet pooling of this method is illustrated in Figure \ref{fig:fopropagation}:

\begin{figure}[h]
\centering
\includegraphics[scale=0.5]{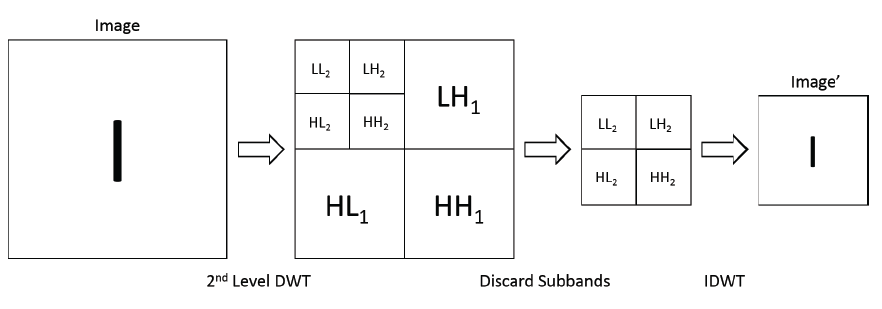}
\caption{Wavelet Pooling Forward Propagation Algorithm  \cite{williams2018wavelet}}
\label{fig:fopropagation}
\end{figure}

The backpropagation of the wavelet pooling is completed by reversing the process of forward propagation. The image feature undergoes 1st order wavelet decomposition first, then a new 1st level decomposition is created by the detail coefficient subbands upsample with a factor of 2. Therefore, the decomposition becomes the 2nd level decomposition. In the end, the image feature is reconstructed with this new 2nd order wavelet decomposition for further backpropagation with the IDWT. The details of the backpropagation algorithm of wavelet pooling are demonstrated Figure \ref{fig:backpagation}:

\begin{figure}[h]
\centering
\includegraphics[scale=0.5]{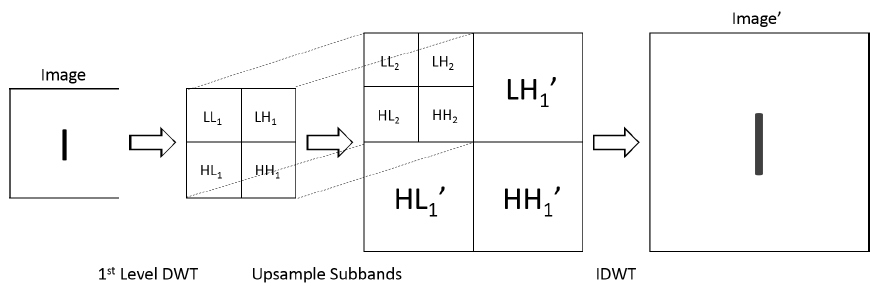}
\caption{Wavelet Pooling Backpropagation Algorithm  \cite{williams2018wavelet}}
\label{fig:backpagation}
\end{figure}

\subsection{Conclusions}

The interests in photorealistic style transfer have been growing driven by the development of gaming and AR/VR. In this review, we presented a comprehensive view of the development of style transfer, including photorealistic style transfer, with detailed discussions on several classic articles in this area. Although the booming of deep learning algorithms during the past several years has significantly improved the development of photorealistic style transfer, the deep learning algorithms only achieved the best photorealism with the help of traditional image processing techniques such as wavelet transform. Therefore, in this review, the contributions of the traditional image processing techniques on feature extraction were discussed as well.   

\EOD
\printbibliography

\end{document}